\documentclass[sigconf]{acmart}
\usepackage{url}
\usepackage[T1]{fontenc}
\usepackage{verbatim}
\usepackage{amsmath}
    \DeclareMathOperator*{\argmax}{argmax}
\usepackage{multirow}
\usepackage{booktabs}
\usepackage{graphicx}
\usepackage{subfigure}
\usepackage{algorithm}
\usepackage{algpseudocodex}
\usepackage{tabu}
\usepackage{tikz}
\usetikzlibrary{arrows,positioning,shapes.multipart,patterns} 
\tikzset{
	>=stealth',
	punkt/.style={
		rectangle,
		rounded corners,
		draw=black, thick,
		text centered,
            align=center,
		font=\footnotesize,
	},
	rect/.style={
		rectangle,
		draw=black, thick,
		text centered,
		font=\footnotesize,
	},
	pil/.style={
		->,
		thick,
		shorten <=2pt,
		shorten >=2pt,
		font=\footnotesize,
	},
}
\usepackage{tkz-graph}
\usepackage{float}
\usepackage{adjustbox}
\usepackage{dashrule}
\renewcommand{\Pr}{\mathrm{Pr}}

\AtBeginDocument{%
  \providecommand\BibTeX{{%
    \normalfont B\kern-0.5em{\scshape i\kern-0.25em b}\kern-0.8em\TeX}}}

\setcopyright{acmcopyright}
\copyrightyear{2024}
\acmYear{2024}
\acmDOI{XXXXXXX.XXXXXXX}

\acmConference[TheWebConf '24]{The ACM Web Conference}{May 13--17, 2024}{Singapore} 
\begin{document}

\title{Enhancing Complex Question Answering over Knowledge Graphs through Evidence Pattern Retrieval}

\author{Wentao Ding}
\affiliation{
   \institution{State Key Laboratory for Novel
Software Technology, Nanjing University}
   \country{China}
}
\orcid{0000-0002-7238-7115}
\email{wtding@smail.nju.edu.cn}

\author{Jinmao Li}
\affiliation{
   \institution{State Key Laboratory for Novel
Software Technology, Nanjing University}
   \country{China}
 }\email{jmli@smail.nju.edu.cn}

\author{Liangchuan Luo}
\affiliation{
   \institution{State Key Laboratory for Novel
Software Technology, Nanjing University}
   \country{China}
 }\email{liangchuanluo@smail.nju.edu.cn}

\author{Yuzhong Qu}
\affiliation{
   \institution{State Key Laboratory for Novel
Software Technology, Nanjing University}
   \country{China}
 }
\orcid{0000-0003-2777-8149}
\email{yzqu@nju.edu.cn}


\begin{abstract}
Information retrieval (IR) methods for KGQA consist of two stages: subgraph extraction and answer reasoning. We argue current subgraph extraction methods underestimate the importance of structural dependencies among evidence facts. 
We propose Evidence Pattern Retrieval (EPR) to explicitly model the structural dependencies during subgraph extraction.
We implement EPR by indexing the atomic adjacency pattern of resource pairs. Given a question, we perform dense retrieval to obtain atomic patterns formed by resource pairs. We then enumerate their combinations to construct candidate evidence patterns. These evidence patterns are scored using a neural model, and the best one is selected to extract a subgraph for downstream answer reasoning.
Experimental results demonstrate that the EPR-based approach has significantly improved the F1 scores of IR-KGQA methods by over 10 points on ComplexWebQuestions and achieves competitive performance on WebQuestionsSP.
\end{abstract}

% \keywords{Do, Not, Us, This, Code, Put, the, Correct, Terms, for,



\maketitle

\section{Introduction}
With the rapid progress of large-scale knowledge graphs (KGs), there is an increasing demand for convenient and precise access to information stored within these KGs. 
Question answering over knowledge graphs (KGQA), i.e., the task of answering factual questions using knowledge graph facts, has gained significant attention~\cite{survey}. 
The mainstream KGQA approaches can be roughly classified into semantic parsing (SP) approaches and information retrieval (IR) approaches. SP approaches parse natural language questions into executable queries and IR approaches retrieve answers through neural models. 
In recent years, KGQA research has focused on solving complex questions that require multi-hop reasoning. SP approaches have made significant progress in solving complex questions~\cite{cbr,tiara,rngkbqa,yu2023decaf}. However, their performance relies on extensive gold question-query pairs. 
Without the labor-intensive annotation of gold queries, these SP approaches may either be untrainable or exhibit a significant decrease in performance.
The development of in-context learning techniques has promoted the practical methods for few-shot KGQA. However, current studies~\cite{wang2023knowledgedriven,li-etal-2023-shot,nie2023codestyle,li2023flexkbqa} indicate that these methods still lag behind the performance of the best SP and IR methods.
IR approaches avoid labor-intensive query annotation by collecting the neighboring information of the topic entities (i.e., entities mentioned by the question).
Because the scale of the entire KG does not support efficient training and retrieval, IR systems first extract a subgraph from the KG and then only process the information on this subgraph.
Therefore, subgraph extraction greatly impacts the performance of IR approaches. 
While there has been some effort in extracting high-quality subgraphs~\cite{pullnet,sr,unikgqa}, the performance of IR approaches still lags far behind that of SP approaches on complex questions.

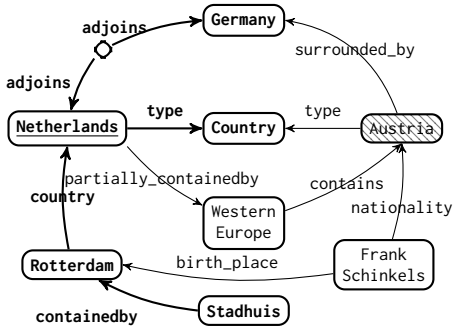
\begin{figure}[tb]
    \centering
    \begin{tikzpicture}
        \node[punkt] (TE1) {\textbf{\texttt{Germany}}};
        \node[punkt, below left = 0.05 and 1.2 of TE1, thick] (B1) {}
            edge[->, bend left = 10, thick] node[pil, left, thick] {\textbf{\texttt{adjoins}}} (TE1.west);
        \node[punkt, below = 1 of TE1] (TE2) {\textbf{\texttt{Country}}};           
        \node[punkt, left = 1 of TE2, thick] (A1) {\underline{\textbf{\texttt{Netherlands}}}}
            edge[->, thick] node[pil, above] {\textbf{\texttt{type}}} (TE2.west)
            edge[<-, bend left = 10, thick] node[pil, left] {\textbf{\texttt{adjoins}}} (B1.south west);
        \node[punkt, right = 1 of TE2, pattern=north west lines, pattern color=black, fill opacity=0.5, text opacity=1] (A2) {\texttt{Austria}}
            edge[->] node[pil, above] {\texttt{type}} (TE2.east)
            edge[->, bend right = 30] node[pil] {\texttt{surrounded\_by}} (TE1.east);
        \node[punkt, below = 2 of TE2] (TE3) {\textbf{\texttt{Stadhuis}}};
        \node[punkt, above left = 0.2 and 1 of TE3, thick] (M1) {\textbf{\texttt{Rotterdam}}}
            edge[<-, bend right = 10, thick] node[pil, below left] {\textbf{\texttt{containedby}}} (TE3.west)
            edge[->, bend left = 10, thick] node[pil] {\textbf{\texttt{country}}} (A1.south);
        \node[punkt, above = 0.6 of TE3] (M2) {\texttt{Western}\\\texttt{Europe}}
            edge[<-, bend left = 10] node[pil] {\texttt{partially\_containedby}} (A1.south east)
            edge[->, bend right = 10] node[pil] {\texttt{contains}} (A2.south);
        \node[punkt, right = 2.8 of M1] (M3) {\texttt{Frank}\\\texttt{Schinkels}}
            edge[->, bend left = 10] node[pil, above] {\texttt{birth\_place}} (M1.east)
            edge[->, bend right = 10] node[pil, below] {\texttt{nationality}} (A2.south);
    \end{tikzpicture}
    \caption{Facts about question ``\textit{What country, containing Stahuis, does Germany border?}''. The evidence facts are \textbf{bolded.}, the node of the correct answer \texttt{Netherlands} is \underline{underlined}, and the noisy answer \texttt{Austria} is shaded. Austria is a noisy answer since it does not contain Stahuis, but the relations on the paths between them express similar meanings and confuse the answer reasoning model.}
    \label{fig:enn}
\end{figure}

We find that current IR studies primarily focus on how to obtain the answer(s) but pay insufficient attention to non-answer parts in the extracted subgraph.
An entity can be considered an answer to the input question only if specific facts surrounding it serve as corresponding evidence. Whether a fact acts as evidence depends not only on its content but also on how it describes the topic entities and answers, specifically the structural dependencies among the relevant facts.
Although current studies have considered iteratively selecting facts with question-related relations during subgraph extraction~\cite{sr,unikgqa} or the downstream reasoning~\cite{nsm-h} stage, their approaches do not provide an explicit semantic representation of the structural dependencies. We find that they sometimes include more noises in the retrieval results which may hurt performance. As illustrated in Figure~\ref{fig:enn}, the facts about the noisy answer \texttt{Austria} have very similar relations to the evidence facts, which may cause ranking errors.

In this paper, we formulate the structural dependencies as \textit{evidence pattern} and propose \textit{evidence pattern retrieval} (EPR) to reduce noises in subgraph extraction. Specifically, evidence pattern models how necessary resources (topic entities and relations) are connected to support a knowledge graph node as an answer to the question. Figure~\ref{fig:efnep} illustrates the corresponding evidence pattern for the question in Figure~\ref{fig:enn}. Section~\ref{sec:def} will introduce the concept of evidence pattern in detail.
We train a neural model to retrieve possible atomic patterns for the input question. We obtain candidate EPs by enumerating combinations of atomic patterns and train a scoring model to select the best EP for subgraph extraction.
We incorporate EPR into existing answer reasoning methods and conduct experiments on ComplexWebQuestions~\cite{cwq} and WebQuestionsSP~\cite{webqsp}, the two most widely used datasets in IR-KGQA evaluations. Experimental results show that evidence pattern retrieval greatly enhances the performance of IR methods on ComplexWebQuestions with competitive performance on WebQuestionsSP.

The rest of this paper is organized as follows. The next section summarizes related work. The third section formulates the IR-KGQA task and EPR. The fourth section introduces the implementation of EPR in detail. The fifth section presents the experimental results with analysis. The last section concludes this paper.

\section{Related Work}
\subsection{KGQA Benchmarks}
The research community has proposed lots of question-answering datasets~\cite{qald,simq,free917,webq} over large-scale open-domain knowledge graphs over the past decades. 
Recently, researchers have begun to pay more attention to complex questions that require reasoning on multi-hop evidence~\cite{cwq,hu-etal-2018-state}. 
Some datasets follow WebQuestions~\cite{webq} to collect questions first and then annotate them.
\citet{bao-etal-2016-constraint} uses question-answer (QA) pairs collected from WebQuestions~\cite{webq} together with manually labeled QA pairs to construct ComplexQuestions.
\citet{cwq} proposes ComplexWebQuestions, where complex questions are generated by composing simpler questions in WebQuestionsSP~\cite{webqsp}, the cleaned version of WebQuestions that is rephrased by AMT workers. 
Some other studies have taken a different approach, generating queries first and then providing corresponding natural language questions for those queries.
MetaQA~\cite{metaqa} and LC-QuAD~\cite{lcquad} generate questions with several tens of pre-defined templates. LC-QuAD 2.0~\cite{lcquad2} extends the framework of LC-QuAD via revised templates and crowd-sourcing tasks.
\citet{grailqa} constructs GrailQA with crowd-powered paraphrasing on manually annotated canonical questions for evaluating KGQA in three different levels of generalization.
\citet{kqapro} introduces KQA Pro with a compositional programming language KoPL to represent the reasoning process explicitly.

\begin{figure}[tb]
    \centering
    \subfigure[Evidence facts]{
    \begin{tikzpicture}[x=25]
        \node[punkt] (TE1) {\texttt{Germany}};
        \node[punkt, below left = 0.05 and 1.5 of TE1] (B1) {}
            edge[->, bend left = 10] node[pil, left] {\texttt{adjoins}} (TE1.west);
        \node[punkt, below = 1 of TE1] (TE2) {\texttt{Country}};           
        \node[punkt, left = 1 of TE2] (A1) {\texttt{Netherlands}}
            edge[->] node[pil, above] {\texttt{type}} (TE2.west)
            edge[<-, bend left = 10] node[pil, left] {\texttt{adjoins}} (B1.south west);
        \node[punkt, below = 1.5 of TE2] (TE3) {\texttt{Stadhuis}};
        \node[punkt, above left = 0.2 and 1 of TE3] (M1) {\texttt{Rotterdam}}
            edge[<-, bend right = 10] node[pil, below left] {\texttt{containedby}} (TE3.west)
            edge[->, bend left = 10] node[pil] {\texttt{country}} (A1.south);
    \end{tikzpicture}
    }
    \subfigure[Evidence pattern]{
    \begin{tikzpicture}[x=25]
        \node[punkt] (TE1) {\texttt{Germany}};
        \node[punkt, below left = 0.05 and 1.5 of TE1] (B1) {$?x_1$}
            edge[->, bend left = 10] node[pil, left] {\texttt{adjoins}} (TE1.west);
        \node[punkt, below = 1 of TE1] (TE2) {\texttt{Country}};           
        \node[punkt, left = 1.8 of TE2] (A1) {$?x_2$}
            edge[->] node[pil, above] {\texttt{type}} (TE2.west)
            edge[<-, bend left = 10] node[pil, left] {\texttt{adjoins}} (B1.south);
        \node[punkt, below = 1.5 of TE2] (TE3) {\texttt{Stadhuis}};
        \node[punkt, above left = 0.2 and 1 of TE3] (M1) {$?x_3$}
            edge[<-, bend right = 10] node[pil, below left] {\texttt{containedby}} (TE3.west)
            edge[->, bend left = 10] node[pil] {\texttt{country}} (A1.south);
    \end{tikzpicture}
    }
    \caption{The evidence facts (a) of question ``\textit{What country, containing Stahuis, does Germany border?}'' and the corresponding pattern (b).}
    \label{fig:efnep}
\end{figure}
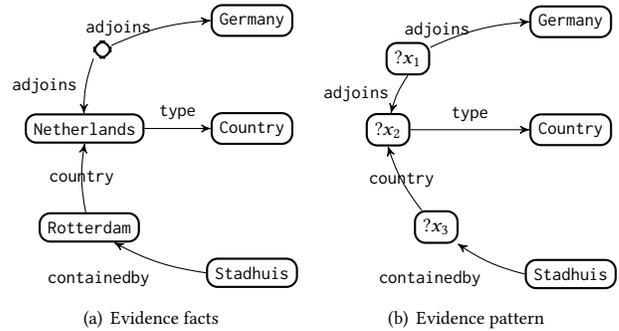

\subsection{Information Retrieval Methods for KGQA}
The mainstream solutions of KGQA first find the entities in the question (i.e., topic entities), and then search for the answers around these entities~\cite{survey}.
Information retrieval methods for KGQA (IR-KGQA methods) use neural models to directly score candidate answers and determine an answer set based on a score threshold. 
The earlier IR-KGQA methods mainly focus on simple questions that only require a 1-hop reasoning~\cite{simq,dong-etal-2015-question,xu-etal-2016-question,kvmem}. For complex questions, IR-KGQA methods limit the search space by considering a subgraph of the entire KG. \citet{embedded_kgqa} models the QA task via link prediction models. It restricts KG to 2-hops neighbors of topic entities and prunes the relations according to the training data.
GraftNet~\cite{graftnet} heuristically extracts a subgraph with personalized PageRank scores computed on the neighborhoods of topic entities. It ranks the extracted nodes via a graph convolutional network to predict the answers. Recent studies learn to extract question-specific subgraphs through neural models. PullNet~\cite{pullnet} proposes a framework to iteratively construct question-specific subgraphs. It trains a GCN to identify subgraph nodes that should be ``pulled'' and predict the answers following GraftNet.
\citet{nsm-h} enhances neural state machine with a teacher network for providing intermediate supervision signals.
\citet{sr} proposes a trainable subgraph retriever to reduce the reasoning bias. It expands relation paths via a sequential decision process. 
UniKGQA~\cite{unikgqa} unifies subgraph extraction and answer reasoning via a semantic matching module for matching question-related relations and a propagation module to propagate the matching information.

The recent advancements in subgraph extraction have highlighted the significance of extracting question-related facts. However, these efforts primarily concentrate on specifying individual facts or relations, neglecting the crucial structural dependencies that enable these facts to support the answers.

\subsection{Semantic Parsing Methods for KGQA}
Semantic parsing methods for KGQA (SP-KGQA methods), distinct from IR-KGQA methods, represent another major category of mainstream methods that parse questions into executable queries to obtain the answer(s). Classic SP-KGQA methods depend on syntactic parsing of questions, which can be challenging when dealing with heterogeneity between questions and query expressions~\cite{survey}. Current representative SP-KGQA methods utilize pre-trained Seq2Seq models to generate queries and use retrieved information (e.g., question-relevant context or intermediate results) to augment the input~\cite{cbr,rngkbqa,yu2023decaf} or restrict the decoding space~\cite{hu-etal-2022-logical,tiara,gu-etal-2023-dont}. For examples, \citet{cbr} proposes the idea of case-based reasoning to leverage query structures from similar questions, significantly improving performance on complex questions. \citet{rngkbqa} uses heuristically retrieved candidate queries as auxiliary information to augment the generation. \citet{tiara} proposes multi-grained retrieval to restrict the decoding process with relevant KG context. These methods require fine-tuning Seq2Seq models with gold query annotations. \citet{yu2023decaf} proposes a joint decoding approach that can work even in situations with only answer annotations but with a noticeable drop in performance. \citet{cao-etal-2022-program} leverages the annotation on the rich-resourced scenarios to improve the performance on scenarios that lack query annotations. 

While SP-KGQA methods perform well on the benchmarks, their performance heavily relies on gold query annotations.
The recent advancement in in-context learning enables practical few-shot KGQA methods that only require a limited number of query annotations. However, their strict few-shot performance still lags behind the current SOTA methods~\cite{li-etal-2023-shot,nie2023codestyle,li2023flexkbqa}.

\section{Task Formulation}~\label{sec:def}
In this paper, we formalize a knowledge graph (KG) $\mathcal{G}$  as a set of triplet facts to describe entities $E$ via their relations $R$, i.e., $\mathcal{G} \subseteq E \times R \times E$.
For a given question $q$, the KGQA task is to obtain the answer(s) $A_q \subseteq E$ according to $\mathcal{G}$. 
The information retrieval (IR) KGQA methods maximize the probability $\Pr(e \in A_q)$ to distinguish $A_q$ from other entities.
Since exploring the entire KG is computationally expensive, the majority of practical IR-KGQA methods operate under the assumption that a question-relevant subgraph $SG_q^* \subseteq \mathcal{G}$ exists, where $\Pr(e \in A_q | q, \mathcal{G}) = \Pr(e \in A_q | q, SG_q^*)$. Therefore, IR-KGQA methods are divided into two stages in practice.
The first stage extracts a question-relevant subgraph to approximate $SG_q^* \subseteq \mathcal{G}$, and the second stage maximizes the probability of answers. We denote them as \textit{subgraph extraction} and \textit{answer reasoning} respectively.
From the probabilistic perspective, subgraph extraction models a latent distribution $\Pr_{\phi}$ and maximizes $\Pr_{\phi}(SG_q^*)$, answer reasoning models a distribution $\Pr_{\psi}$ on the extracted subgraph to approximate $\Pr$. We formulate them as follows:
\begin{equation}\begin{split}
& SG_q = \textsc{Ext}(\mathcal{G}, q, \Pr_{\phi}),\\
& A_{pred} = \left\{e\in SG_q \mid \Pr_{\psi}(e \in A_q | q, SG_q) > \theta \right\},
\end{split}\end{equation}
where \textsc{Ext} denotes a subgraph extractor, $A_{pred}$ denotes the predicted answers, $\theta$ is a confidence threshold for determining the answer set.

This paper focuses on subgraph extraction. We argue that including noisy facts in subgraph extraction will affect the consequent reasoning stage, i.e., the ideal subgraph $SG^*_q$ for question $q$ is formed by a minimal set of \textit{evidence facts}. Given an appropriate similarity measure $sim$ over graphs and questions, $SG^*_q$ should be the most similar subgraph to question $q$, i.e.,
\begin{equation}
    \argmax_{SG \subset \mathcal{G}} \Pr_{\phi}(SG|q) = \argmax_{SG \subset \mathcal{G}} sim(SG, q).
\end{equation}
We assume that the similarity is determined by the adjacency structure over topic entities. We model it as the \textit{evidence pattern} $pat(SG, q)$ of $SG$. Specifically, the evidence pattern is a variable substitution of $SG$, where all entities not appearing in the question $q$ are replaced by variable symbols, as illustrated in Figure~\ref{fig:efnep}. 
% We extend the similarity measure to evidence patterns and assume that the similarity of any subgraph $SG$ to question $q$ equals the similarity of its evidence pattern to $q$, i.e., $sim(SG,q) = sim(pat(SG),q)$.
Therefore, the task of subgraph extraction can be achieved through \textit{evidence pattern retrieval}. The extraction target can be formulated as follows:
\begin{equation}
\argmax_{SG \subset \mathcal{G}} sim(pat(SG, q), q).
\end{equation}
Specifically, after retrieving the most appropriate evidence pattern $P$, we instantiate $P$ by the maximum graph that satisfies $P$, i.e., the subgraph with maximum entities. This subgraph of KG will be provided to the consequent answer reasoning. 

\section{Evidence Pattern Retrieval}
\begin{figure}[ht]
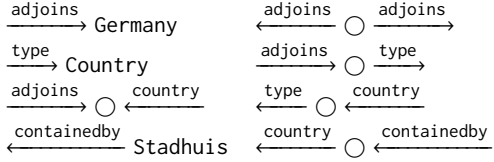

    \centering
    \begin{tabular}{ll}
        $\xrightarrow{\texttt{adjoins}} \texttt{Germany}$ & $ \xleftarrow {\texttt{adjoins}} \bigcirc \xrightarrow{\texttt{adjoins}}$ \\
        $\xrightarrow{\texttt{type}} \texttt{Country}$ & $\xrightarrow{\texttt{adjoins}} \bigcirc \xrightarrow{\texttt{type}}$ \\
        $\xrightarrow{\texttt{adjoins}} \bigcirc \xleftarrow{\texttt{country}}$ & $\xleftarrow{\texttt{type}} \bigcirc \xleftarrow{\texttt{country}}$ \\
        $\xleftarrow{\texttt{containedby}} \texttt{Stadhuis}$ & $\xleftarrow{\texttt{country}} \bigcirc \xleftarrow{\texttt{containedby}}$ \\
    \end{tabular}%
    \caption{Atomic patterns appeared on the evidence pattern in Figure~\ref{fig:efnep}.}
    \label{fig:cp}
\end{figure}

For a specific knowledge graph $\mathcal{G}$. The retrieval space can be denoted as $\left\{pat(SG) \mid SG \subseteq \mathcal{G} \right\}$. 
Given that the space exceeds the scale of manageable storage, our approach analyzes EP at the granularity of the atomic adjacency structure among entities and relations, denoted as \textit{atomic patterns} (APs).
Each AP consists of a pair of adjacent resources and defines their connection structure. Specifically, AP includes the directed connection of entity-relation pairs (denoted as ER-APs) and relation-relation pairs (denoted as RR-APs). As illustrated in Figure~\ref{fig:cp}, the EP of a question can be covered by corresponding APs. 
We achieve EPR through the indexing and retrieval of atomic patterns, encompassing all conceivable EP instances.

For an input question, we use a fine-tuned bi-encoder model to retrieve candidate APs and enumerate the possible EPs with an iterative pattern expansion algorithm. If there are multiple candidate EPs, we score them via a ranking model to select the best EP.

\subsection{Atomic Pattern Retrieval}
For an input question, the space of possible ER-APs is restricted by the topic entities, but the space of possible RR-APs includes all adjacent relation pairs in KG. Therefore, we have to build a fast index for retrieving candidate RR-APs.

We follow the dense retrieval fashion~\cite{dpr} to build Faiss~\cite{faiss} indexes of RR-APs in the given KG. We encode the input question and the RR-APs via two independent BERT~\cite{devlin-etal-2019-bert} models. Given a question $q$ and an RR-AP $p$, we compute the dot similarity of the embeddings of corresponding \texttt{[CLS]} tokens, i.e.,
\begin{equation}\begin{split}
& V_q = \textsc{BertCLS}_1(q),\\
& V_p = \textsc{BertCLS}_2(p),\\
& \mathrm{sim}(p,q) = (V_q)^T \cdot V_p,
\end{split}\end{equation}
where \textsc{BertCLS} denotes the representation of corresponding \texttt{[CLS]}. Specifically, we serialize each RR-AP via the corresponding relation labels and a link tag for denoting how the relations are connected. Table~\ref{tab:scp} illustrated how to serialize RR-APs of different structures.

\begin{table}[ht]
    \centering
    \caption{Serialization of RR-APs.}
    \label{tab:scp}
    \begin{tabular}{cc}
        \toprule
        \textbf{RR-APs} & \textbf{Serialization}  \\
        \midrule
        $\xleftarrow{\texttt{rel$_1$}} \bigcirc \xrightarrow{\texttt{rel$_2$}}$ &  \texttt{[CLS] rel$_1$ SS rel$_2$ [SEP]} \\
        $\xleftarrow{\texttt{rel$_1$}} \bigcirc \xleftarrow{\texttt{rel$_2$}}$ &  \texttt{[CLS] rel$_1$ SO rel$_2$ [SEP]} \\
        $\xrightarrow{\texttt{rel$_1$}} \bigcirc \xrightarrow{\texttt{rel$_2$}}$ &  \texttt{[CLS] rel$_1$ OS rel$_2$ [SEP]} \\
        $\xrightarrow{\texttt{rel$_1$}} \bigcirc \xleftarrow{\texttt{rel$_2$}}$ &  \texttt{[CLS] rel$_1$ OO rel$_2$ [SEP]} \\
        \bottomrule
    \end{tabular}
\end{table}

The dense encoders are trained with the cross-entropy loss on the output logits. 
We use heuristically constructed pseudo EPs to avoid manual annotation. For each question in the training set, we randomly select one of its answers, and subsequently collect 1 or 2 hops paths between the topic entities and the selected answer to construct a pseudo-EP.\footnote{CVT connections over entities in Freebase are treated as 1-hop during the collection process. Therefore, the maximum length of relation paths is 4, rather than 2.}. The RR-APs on the pseudo-EP(s) are considered positive samples. The negative samples are generated in two ways. Half of the negative samples are generated by randomly replacing one relation or the tag of a positive sample, and the others are randomly sampled over all the RR-APs collected from the KG.

In the test process, we use the dense encoders to retrieve $K$ most relevant RR-APs and collect all the ER-APs of topic entities as candidate APs. 

\begin{algorithm}[tbh]
    \caption{The construction of candidate EPs}
    \label{alg:eep}
    \begin{algorithmic}[1]
        \Function{Enumerate}{$\tau, AP_{ER}, AP_{RR}$}
            \State $C \leftarrow \emptyset$
            \For {$p \in AP_{ER}$}
                \State $AP'_{ER} \leftarrow AP_{ER} \setminus \left\{q \in AP_{ER} \mid q.ent = p.ent \right\}$ \label{line:te1}
                \State $C \leftarrow C \cup \textsc{IterExpand}(p, \tau, AP'_{ER}, AP_{RR})$
            \EndFor
            \State \Return {$C$}
        \EndFunction
    \end{algorithmic}
\end{algorithm}

\begin{algorithm}[tbh]
    \caption{The iterative expansion of under-construction EPs}
    \label{alg:expd}
    \begin{algorithmic}[1]
        \Function{IterExpand}{$P, \tau, AP_{ER}, AP_{RR}$}
            \State $C \leftarrow \emptyset$
            \If {$\textsc{IsValidPat}(P)$} \label{line:valid}
                \State $C \leftarrow C \cup \{P\}$
            \EndIf
            \If {$\left|P\right| = \tau$}
                \State \Return {$C$}
            \EndIf
            \For {$p \in \left\{p \in AP_{ER} \mid \textsc{Expandable}(P, p)\right\}$} \label{line:expand1}
                \State $P' \leftarrow \textsc{Expand}(P, p)$
                \State $AP'_{ER} \leftarrow AP_{ER} \setminus \left\{q \in AP_{ER} \mid q.ent = p.ent \right\}$ \label{line:te2}
                \State $C' \leftarrow \textsc{IterExpand}(P', \tau, AP'_{ER}, AP_{RR})$
                \State $C \leftarrow C \cup C'$
            \EndFor
            \For {$p \in \left\{p \in AP_{RR} \mid \textsc{Expandable}(P, p))\right\}$} \label{line:expand2}
                \State $P' \leftarrow \textsc{Expand}(P, p)$
                \State $C' \leftarrow \textsc{IterExpand}(P', \tau, AP_{ER}, AP_{RR})$
                \State $C \leftarrow C \cup C'$
            \EndFor
            \State \Return {$C$}
        \EndFunction
    \end{algorithmic}
\end{algorithm}

\begin{figure}[tbh]
    \centering
    \begin{tikzpicture}[x=20]
        \node[punkt] (TE1) {\texttt{Germany}};
        \node[punkt, below left = 0.05 and 1.5 of TE1, color = teal] (B1) {$?x_1$}
            edge[->, bend left = 10] node[pil, left] {\texttt{adjoins}} (TE1.west);
        \node[punkt, below = 1 of TE1, color = blue] (TE2) {$?x_3$};           
        \node[punkt, left = 1.8 of TE2, color = teal] (A1) {\color{blue} $?x_2$}
            edge[->, color = blue] node[pil, above] {\texttt{type}} (TE2.west)
            edge[<-, bend left = 10, color = teal] node[pil, left] {\texttt{adjoins}} (B1.south);
       
        \node[below left = 2 and 0.4 of A1] (LSTART) {}
            edge[<-, bend left = 15, shorten <= 10, shorten >= 10] node[left, font=\scriptsize, align = center] {Expand ER-AP\\ \color{blue} $\xleftarrow{\texttt{type}} \texttt{Country}$} (A1.south);
        \node[punkt, below right = 0 and 0 of LSTART] (LTE1) {\texttt{Germany}};
        \node[punkt, below left = 0.05 and 1.5 of LTE1] (LB1) {$?x_1$}
            edge[->, bend left = 10] node[pil, left] {\texttt{adjoins}} (LTE1.west);
        \node[punkt, below = 1 of LTE1, color = blue] (LTE2) {\texttt{Country}};           
        \node[punkt, left = 1.8 of LTE2, color = blue] (LA1) {$?x_2$}
            edge[->, color = blue] node[pil, above] {\texttt{type}} (LTE2.west)
            edge[<-, bend left = 10] node[pil, left] {\texttt{adjoins}} (LB1.south);

        \node[below right = 2 and 0.6 of TE2] (RSTART) {}
            edge[<-, bend right = 15, cap=round, shorten <= 10, shorten >= 10] node[right, font=\scriptsize, align = center] {Expand RR-AP\\ \color{teal} $\xrightarrow{\texttt{adjoins}} \bigcirc \xleftarrow{\texttt{country}}$} (TE2.south);
        \node[punkt, below right= 0 and 1 of RSTART] (RTE1) {\texttt{Germany}};
        \node[punkt, below left = 0.05 and 1.5 of RTE1, color = teal] (RB1) {$?x_1$}
            edge[->, bend left = 10] node[pil, left] {\texttt{adjoins}} (RTE1.west);
        \node[punkt, below = 1 of RTE1] (RTE2) {$?x_3$};           
        \node[punkt, left = 1.8 of RTE2, color = teal] (RA1) {$?x_2$}
            edge[->] node[pil, above] {\texttt{type}} (RTE2.west)
            edge[<-, bend left = 10, color = teal] node[pil, left] {\texttt{adjoins}} (RB1.south);
        \node[punkt, below left = 0.7 and 1 of RTE2, color = teal] (RM1) {$?x_4$}
            edge[->, bend left = 10, color = teal] node[pil] {\texttt{country}} (RA1.south);
    \end{tikzpicture}
    \caption{Expand an under-constructed EP with an ER-AP (the left side) or an RR-AP (the right side).}
    \label{fig:expd}
\end{figure}
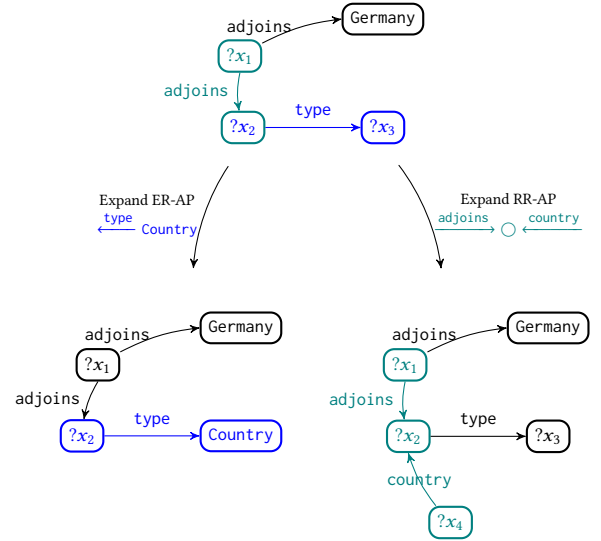

\subsection{Candidate Evidence Pattern Construction}
After retrieving the APs, we enumerate all possible EPs with Algorithm~\ref{alg:eep}. 
The algorithm starts from an ER-AP and iteratively expands the under-construction EP by Algorithm~\ref{alg:expd} \textsc{IterExapnd}. 
\textsc{IterExapnd} uses a threshold $\tau$ to control the maximum size of EPs. The value of $\tau$ varies with different datasets. 
Since APs have already recorded the atomic adjacency, the algorithm only needs to check whether the adjacency recorded by an AP is consistent with the current EP. An atomic expansion happens on a variable node. The variable will be replaced by a topic entity (for expansion with ER-AP) or extended with a relation (for expansion with RR-AP), as illustrated in Figure~\ref{fig:expd}.

Specifically, the algorithm assumes that each topic entity can only be expanded once (line~\ref{line:te1} of Algorithm~\ref{alg:eep} and line~\ref{line:te2} of Algorithm~\ref{alg:expd}), but the use of RR-APs is not limited. After expansion, the EPs that can provide answers will be considered as candidates (line~\ref{line:valid} of Algorithm~\ref{alg:expd}).
There are two critical predictive functions, \textsc{IsValidPat} and \textsc{Expandable}, with the following criteria for their checks:

\textsc{IsValidPat} checks whether the pattern could correspond to the meaning of the question. It requires the pattern to include all topic entities and also demands the existence of knowledge graph nodes that satisfy the variables in the pattern. 
To ensure the simplicity of the pattern and avoid introducing meaningless relations, we do not accept an arbitrary expansion of relation paths. Specifically, only two types of structures are considered valid:
\begin{itemize}
    \item All endpoints of the pattern (i.e., nodes with a degree of 1) are topic entities. In this case, the evidence pattern is a minimal structure that can connect all the topic entities.
    \item There is only one variable endpoint. We allow this exception in case the correct pattern is a simple triplet or all topic entities describe the query target through a common intermediate.
\end{itemize}

\textsc{Expandable} assesses whether a pattern $P$ can be expanded with atomic pattern $p$. The \textsc{Expand} process will try to expand $P$ with $p$ to obtain a new EP $P'$. \textsc{Expandable} return true if only it is possible to construct an expanded $P'$ where all corresponding atomic patterns appear in $AP_{ER}, AP_{RR}$. Taking the expansions in Figure~\ref{fig:expd} as an example, the expansion with ``$\xrightarrow{\texttt{adjoins}} \bigcirc \xleftarrow{\texttt{country}}$'' requires the existence of ``$\xleftarrow{\texttt{type}} \bigcirc \xleftarrow{\texttt{country}}$''.

\subsection{Evidence Pattern Ranking}
The enumeration will generate multiple candidate EPs, and we model the probability $\Pr_{\phi}$ over candidate EPs via a BERT-implemented cross-encoder. The model design is similar to the query ranking models of the mainstream semantic parsing methods~\cite{grailqa,rngkbqa}.
For a question $q$ and a candidate EP $P$, the process can be formulated as follows:
\begin{equation}
sim(P,q)=\textsc{Linear}(\textsc{BertCLS}([q;P])).
\end{equation}
We serialize the candidates as sequences of triplets, where resources are denoted by their Freebase labels (including domains of relations). We concatenate the serialization of EP and the corresponding question as the input of the cross-encoder. 
Sample input for the EP illustrated in Figure~\ref{fig:efnep} is demonstrated as follows:
\begin{displaymath}\begin{split}
&\texttt{[CLS] what }\dots\texttt{ border ? [SEP] ?u }\dots\texttt{adjoins ?v ; } \dots \\
&\texttt{?w }\dots\texttt{containedby stadhuis ; ?u }\dots\texttt{type country ; [SEP]}
\end{split}\end{displaymath}

The model is trained with the cross-entropy loss on the output logits during the training process. we take the candidate EPs that cover the most answers as positive examples, and the others as negative examples.

\begin{figure*}[thb]
    \centering
    \subfigure[CWQ]{
        \includegraphics[width=0.45\textwidth]{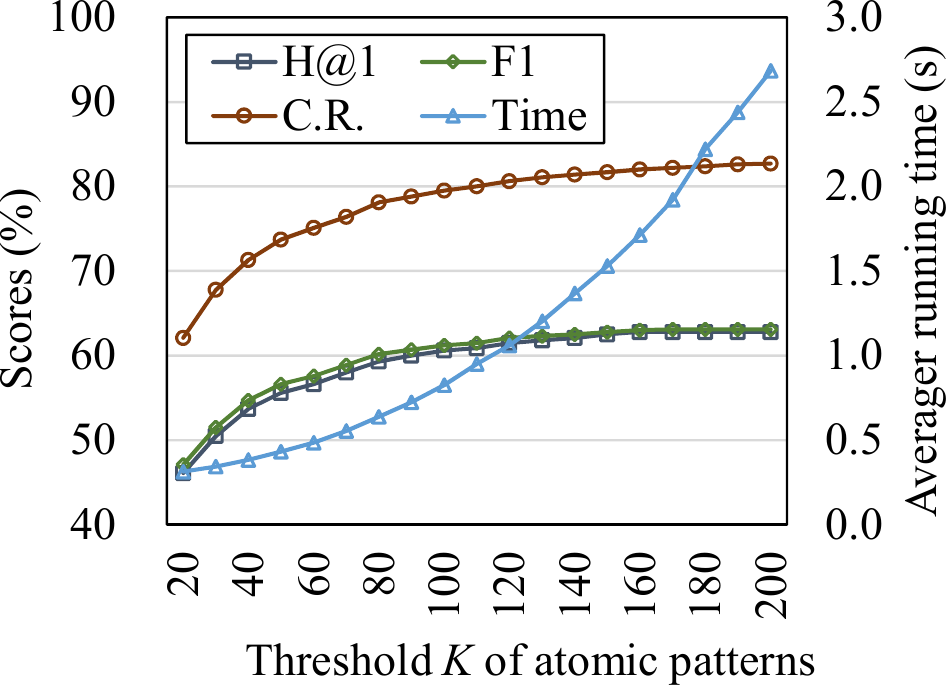}
    }
    \hfill
    \subfigure[WebQSP]{
        \includegraphics[width=0.45\textwidth]{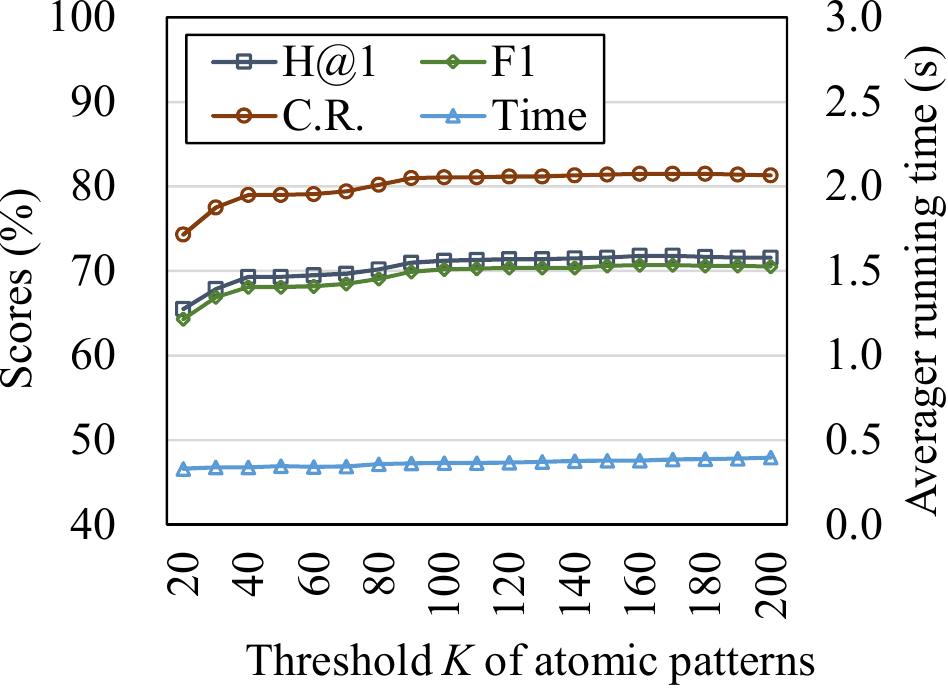}
    }
    \caption{The performance and execution time of EPR+NSM with various numbers of APs on CWQ~(a) and WebQSP~(b).}
    \label{fig:apt}
\end{figure*}

\section{Evaluation}
\subsection{Experimental Settings}
\subsubsection{Datasets}
We evaluate our method on two widely used benchmarks over Freebase~\cite{freebase}, Complex WebQuestions 1.1 (denoted as CWQ) and WebquestionSP (denoted as WebQSP). The statistics about them are presented in Table~\ref{tab:stats}. The split of validation and training data of WebQSP follows \citet{sr}. We use the latest dump of Freebase\footnote{\url{https://developers.google.com/freebase}}. We build a Faiss index with 2,366,590 relation-relation atomic patterns. For the training of the atomic pattern retriever, We generated 19 negative samples for each positive sample, i.e., 5\% of the samples are positive. For the training of evidence pattern ranker, the maximum pos.-to-neg. ratio limit is 1:100. Any additional negative examples will be discarded.

\begin{table}[hbt]
    \centering
    \caption{Statistics of the number of questions.}
    \label{tab:stats}
    \begin{tabu} to 0.9\columnwidth {X[1, l] X[1, r] X[1, r] X[1, r]}
        \toprule
        \rowfont\bfseries Dataset & \#Train & \#Val. & \#Test  \\
        \midrule
        CWQ & 27,639 & 3,519 & 3,531\\
        WebQSP & 2,848 & 250 & 1,639\\
        \bottomrule
    \end{tabu}
\end{table}

\subsubsection{Compared Methods}
We compared with seven two-staged IR-KGQA methods~\cite{embedded_kgqa,kvmem,graftnet,pullnet,nsm-h,sr,unikgqa} and an in-context learning method \textbf{KD-COT}~\cite{wang2023knowledgedriven}. The two-staged IR-KGQA methods propose five subgraph extraction methods and six answering reasoning methods.
The subgraph extraction methods are listed as follows:
\begin{itemize}
    \item \citet{embedded_kgqa} proposes a \textit{relation-pruning strategy} to extract subgraphs connected by allowed relations. We denote it as \textbf{R-Prune}.
    \item \textbf{PPR} denotes the heuristic idea proposed by \citet{graftnet}. They extract a subgraph with \textit{personalized PageRank} scores computed on the neighborhoods of topic entities.
    \item \textbf{PullNet}~\cite{pullnet} iteratively constructs question-specific subgraph by ``pull'' operations on KG and text corpus.   
    \item \textbf{SR} denotes the \textit{subgraph retriver} proposed by \citet{sr}. It expands relation paths via a sequential decision process.
    \item \textbf{UniKGQA}~\cite{unikgqa} unifies subgraph extraction and answer reasoning by computing and propagating matching information between questions and relations.
\end{itemize}
The answer reasoning methods are listed as follows:
\begin{itemize}
    \item \citet{kvmem} proposes a \textit{key-value memory network} to store KG facts, which implicitly models 1-hop neighboring graphs of topic entities. We denoted it as \textbf{KV-Mem}.
    \item \textbf{EmbedKGQA}~\cite{embedded_kgqa} formulates answer reasoning as a link prediction task.
    \item \textbf{GCN} denotes the idea to identify answers via \textit{graph convolutional network}~\cite{graftnet}.
    \item \textbf{NSM} denotes \textit{neural state machine} for KGQA~\cite{nsm-h}. It iteratively generates instruction vectors and updates the entity distribution to predict the final answer(s).
    \item \textbf{UniKGQA}~\cite{unikgqa} uses the same architecture for subgraph extraction and answer reasoning.
    \item \textbf{KD-COT}~\cite{wang2023knowledgedriven} is an in-context learning method. It proposes the \textit{knowledge driven chain-of-thought} reasoning process to iteratively retrieve KG.
\end{itemize}
We use NSM as the answer reasoner to implement our KGQA system.

In addition to the above IR-KGQA methods, we also provide the results of three recent semantic parsing methods for reference, including two fine-tuning methods \textbf{DecAF}~\cite{yu2023decaf} and \textbf{Program Transfer}~\cite{cao-etal-2022-program} and a few-shot method \textbf{KB-Coder}~\cite{nie2023codestyle}. We report the results of these methods results because they can run with and without full annotation of gold queries.

\subsubsection{Implementation Details}
We implemented EPR with Python 3.7 and PyTorch 1.9. The results are obtained on a server with Intel Xeon Gold 5222 CPUs and NVIDIA RTX 3090 GPUs~\footnote{The implementation is available at \url{https://github.com/nju-websoft/EPR-KGQA}.}. We use \texttt{bert-base-uncased} as the neural model for atomic pattern retrieval and evidence pattern ranking. The hyper-parameters are presented in Table~\ref{tab:hp}.

\begin{table}[hbt]
    \centering
    \caption{Hyper-parameters for the models.}
    \label{tab:hp}
    \begin{tabular}{lrr}
        \toprule
        \textbf{Parameters} & \textbf{AP retrieval} & \textbf{EP ranking} \\
        \midrule
        Vector dimension & 768 & 768 \\
        Batch size & 16 & 2 \\
        Epochs & 5 & 10 \\
        Initial learning rate & 2e-5 & 1e-5 \\
        \bottomrule
    \end{tabular}
\end{table}

The size threshold $\tau$ of evidence patterns is decided by the sizes of collected pseudo EPs. Specifically, the thresholds for CWQ and WebQSP are set to 5 and 3 respectively.
We grid-search the threshold $K$ for the maximum number of retrieved RR-APs from 20 to 200 with a step of 10. 
Larger $K$ may increase the running time while improving the answer cover rates.

\begin{table}[tbh]
    \centering
    \caption{The evaluation results(\%) on CWQ. The best results of IR methods are in \textbf{bold}, and the second-best results are \underline{underlined}. $\dagger$ denotes that the method requires gold query annotation of all training questions. $*$ denotes few-shot methods.}
    \label{tab:main}
    \begin{tabular}{lrrrr}
        \toprule
        \multirow{2}{*}{\textbf{Method}} & \multicolumn{2}{c}{\textbf{CWQ}} & \multicolumn{2}{c}{\textbf{WebQSP}} \\
        \cmidrule(lr){2-3} \cmidrule(lr){4-5}
        & H@1 & F1 & H@1 & F1 \\
        \midrule
        \multicolumn{5}{l}{\textit{Semantic Paring Methods}}\\
        \midrule[0.1pt]
        % $^\dagger$CBR-KGQA~\cite{cbr} & - & 70.0 & - & 72.8 \\
        DecAF w/o Gold Query~\cite{yu2023decaf} & 50.5 & - & 74.7 & 49.8 \\
        $^\dagger$DecAF w/ Gold Query~\cite{yu2023decaf} & 68.1 & - & 80.7 & 77.1 \\
        Program Transfer~\cite{cao-etal-2022-program} & 58.1 & 58.7 & 74.6 & 76.5\\
        % $^\ddagger$KB-Binder~\cite{li-etal-2023-shot} & - & - & - & 53.2\\
        % $^\dagger$$^\ddagger$KB-Binder + Retrieval~\cite{li-etal-2023-shot} & - & - & - & 74.4 \\
        $^*$KB-Coder~\cite{nie2023codestyle} & - & - & - & 60.5\\
        $^\dagger$$^*$KB-Coder + Retrieval~\cite{nie2023codestyle} & - & - & - & 75.2 \\
        % $^*$FlexKBQA~\cite{li2023flexkbqa} & - & - & - & 60.6\\
        \midrule    
        \multicolumn{5}{l}{\textit{Information Retrieval Methods}}\\
        \midrule[0.1pt]
        KV-MeM~\cite{kvmem} & 18.4 & 15.7 & 46.6 & 34.5\\
        R-Prune + EmbedKGQA\cite{embedded_kgqa} & 32.0 & - & 66.6 & - \\
        PPR + GCN~\cite{graftnet} & 36.8 & 32.7 & 66.4 & 60.4 \\
        PullNet + GCN~\cite{pullnet}  & 45.9 & - & 68.1 & -\\
        PPR + NSM~\cite{nsm-h} & 47.6 & 42.4 & 68.5 & 62.8\\
        SR + GCN~\cite{sr} & 49.0 & 42.7 & 66.7 & 63.1\\
        SR + NSM~\cite{sr} & 50.2 & 47.1 & 69.5 & 64.1 \\
        UniKGQA + NSM~\cite{unikgqa} & 49.2 & - & 69.1 & - \\
        UniKGQA + UniKGQA~\cite{unikgqa} & 50.7 & \underline{48.0} & \textbf{75.1} & \textbf{70.2} \\
        $^*$KD-COT~\cite{wang2023knowledgedriven} & \underline{55.7} & - & 68.6 & 52.5 \\
        %\tabucline[0.1pt on 3pt]{-}
        % EPR$_{AP=200}$ + NSM & \textbf{62.8} & \textbf{63.1} & \underline{71.6} & \textbf{70.5} \\
        EPR$_{K=100}$ + NSM (OURS) & \textbf{60.6} & \textbf{61.2} & \underline{71.2} & \textbf{70.2} \\
        % EPR$_{AP=20}$ + NSM & 46.1 & 47.1 & 65.5 & 64.3 \\
        \bottomrule
    \end{tabular}
\end{table}

\subsubsection{Evaluation Metrics}
We report Hits@1 (denoted as H@1), F1 score, and answer cover rate (denoted as C.R.) of compared methods. Hits@1 directly evaluates whether the top-1 predicted answer is correct. Because some questions have multiple answers, we also use the F1 score to evaluate the system outputs. The predicted answers of our system are truncated according to the default threshold of the answer reasoner. The computation of C.R. follows \citet{sr}, which is the proportion of questions for which the extracted subgraph contains at least one answer. Specifically, C.R. can be regarded as the H@1 with an oracle answer reasoner. It reflects the performance of subgraph extraction and helps to identify the performance of each module separately.

\subsection{Main Results}
Table~\ref{tab:main} reports the main results of compared methods. Due to space limitations, the table only includes our implementation with the top 100 retrieved RR-APs. The comprehensive set of results of our implementation is illustrated in Figure~\ref{fig:apt} and is discussed in Section~\ref{sec:ap}.

Our implementation has achieved a new SOTA for IR methods on the ComplexWebQuestions (CWQ) dataset. Our implementation shows significant improvements, with a +4.9 increase in H@1 (compared to KD-COT) and a +13.2 increase in F1 score (compared to UniKGQA).
% These improvements further expand to +7.1 in H@1 and +15.1 in F1 when $\#AP=200$ is considered.
On the WebQSP dataset, our implementation exhibits competitive performance compared to the SOTA method UniKGQA. There is a decrease in H@1 by -3.9 points, but the F1 score remains similar. Moreover, our implementation surpasses other NSM-based methods. Specifically, our implementation demonstrates notable improvements, with a +1.7 increase in H@1 and a +6.1 increase in F1 compared to other NSM-based methods.

When considering all methods that do not require complete gold query annotations in the training data, our implementation outperforms the current SOTA method, Program Transfer, on the CWQ dataset. The improvements are evident, with a +2.5 increase in F1. However, the results on the WebQSP dataset are lower, showing a -6.3 decrease in F1.

The result reported in Figure~\ref{fig:apt} demonstrated that the ratio of Hits@1 to answer cover rates (i.e., the performance of NSM) is relatively stable. The specific results are reported in Table~\ref{tab:cr}. 
The average ratios of Hits@1 to C.R. are approximately 76\% and 88\% on CWQ and WebQSP, respectively. For reference, SR+NSM with a coverage rate threshold of 0.8 only achieves Hits@1 scores below 0.4 on CWQ and below 0.65 on WebQSP~\cite{sr}. Despite the varying objectives of different subgraph extraction methods making the ratios not immediately comparable, the significant gap suggests that EPR may offer advantages for downstream reasoning tasks.

\begin{table}[htb]
    \centering
    \caption{The average Hits@1(\%), answer cover rates(\%) and their ratio over different threshold $K$. The last column reports the average ratios with the standard deviations in brackets.}
    \label{tab:cr}
    \begin{tabular}{lrrr}
        \toprule
        \textbf{Dataset} & \textbf{avg. H@1} & \textbf{avg. C.R.} & \textbf{avg. H@1/C.R.} \\ 
        \midrule
        CWQ & 59.1 & 77.9 & 75.9 (6.4e-3) \\
        WebQSP & 70.5 & 80.2 & 87.9 (1.8e-3) \\
        \bottomrule
    \end{tabular}
\end{table}

\subsection{Results with Various Number of Retrieved Atomic Patterns}\label{sec:ap}
\begin{figure}[hbt]
    \centering
    \includegraphics[width=0.45\textwidth]{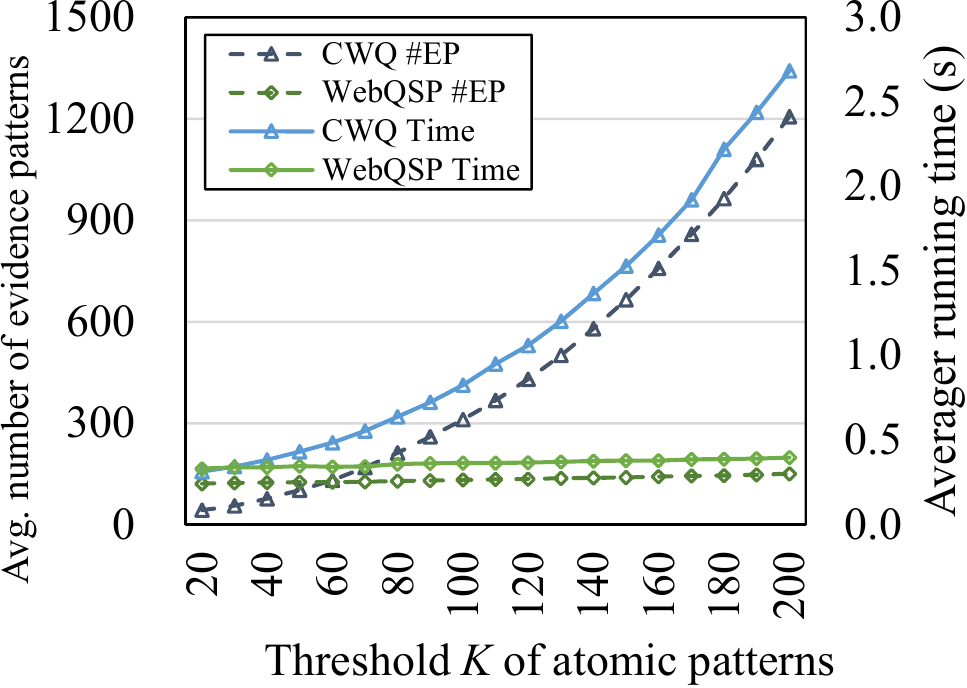}
    \caption{The average running time and number of candidate evidence patterns.}
    \label{fig:time}
\end{figure}

This section discusses the influence of the number of retrieved atomic patterns (APs) on the performance and efficiency of our implementation. 
We computed the average running time and the performance scores under different thresholds of APs, as illustrated in Figure~\ref{fig:apt}.
Our further analysis shows that the increase in running time on CWQ is sorely due to the combinatorial explosion of candidate evidence patterns, as illustrated in Figure~\ref{fig:time}. 
In conclusion, the results indicate a notable impact of the number of APs on complex questions (i.e. CWQ), but the difference it brings is not very significant on relatively simple questions (i.e. WebQSP). Specifically, on CWQ, the range of answer cover rates is from 62.1\% to 82.7\%, and the range of time is from 0.31s to 2.68s. On WebQSP, the range of answer cover rates is from 74.3\% to 81.3\%, and the range of time is from 0.33s to 0.40s.
For complex questions, increases in atomic patterns bring efficiency bottlenecks, but the performance improvement brought by increasing the number is diminishing. For relatively simple questions, the retrieved atomic patterns do not have too many combinations, and the increase in the number threshold has few effects on the system.

\subsection{The Impact of Training Data Size on Pattern Ranking}
We conducted an experiment to assess the influence of training data size on the ranking of candidate evidence patterns. We randomly split the training data into five equal parts and trained the ranking model with various ratios of data, as reported in Table~\ref{tab:epr}. The results indicate that the system's performance is not particularly sensitive to this. We suppose this is because the retrieval of atomic patterns already establishes the question-related KG context, and the ranking model primarily focuses on structural information, with less reliance on data size.

\begin{table}[htb]
    \centering
    \caption{The results (\%) obtained by training the ranking model with various ratios of training data. All results are obtained with $K=100$.}
    \label{tab:epr}
    \begin{tabular}{lrrrrrr}
        \toprule
        \multirow{2}{*}{\textbf{Ratio}} & \multicolumn{3}{c}{\textbf{CWQ}} & \multicolumn{3}{c}{\textbf{WebQSP}} \\
        \cmidrule(lr){2-4} \cmidrule(lr){5-7}
        & C.R. & H@1 & F1 & C.R. & H@1 & F1 \\
        \midrule
        20\% & 77.7 &	59.1 &	59.6 &	75.5 &	65.9 &	64.0  \\
        40\% & 79.4 &	59.8 &	60.6 &	78.5 &	68.3 &	67.8  \\
        60\% & 79.1 &	59.8 &	60.4 &	79.4 &	68.5 &	67.8  \\
        80\% & 78.7 &	59.5 &	60.5 &	81.4 &	70.3 &	69.6  \\
        Full & 79.5 &	60.6 &	61.2 &	81.1 &	71.2 &	70.2  \\
        \bottomrule
    \end{tabular}
\end{table}

\subsection{Error Analysis}
We conduct an error analysis on a sample of 100 questions with incorrect answer predictions obtained from our experiments on CWQ and WebQSP datasets. 
The analysis revealed various issues contributing to the incorrect answer predictions. We classify the main issues into six types and report the statistics in Table~\ref{tab:ea}.
Notably, 20\% of the errors on CWQ and 28\% of the errors on WebQSP stemmed from EPR failing to cover the correct answers. This issue was primarily linked to the insufficiency of retrieved atomic patterns.
A significant proportion of the errors, comprising 54\% on CWQ and 30\% on WebQSP, were caused by non-entity descriptions of the answers. Many of these descriptions involved numerical reasoning, such as expressions like ``higher than 590''. This issue is unlikely to be resolved within the current IR-KGQA framework. Even if we extend evidence patterns to capture numerical patterns, the current answer reasoning methods encounter challenges in handling numerical information.
The construction and ranking of evidence patterns contribute to 10\% of the errors on CWQ and 28\% of the errors on WebQSP.
Only a small proportion of errors on CWQ, 8\%, were caused by imperfect answer reasoning. This observation indicates EPR's effectiveness in reducing the impact of noisy extraction.
Lastly, it is important to note that some of the data has quality issues. For example, our system predicted \texttt{EgyptianArabic} as the top answer to the question ``What kind of language do Egyptians speak?'', which, while seemingly correct, was not part of the annotated answer set.

\begin{table}[htb]
    \centering
    \caption{The statistics about the main issues of incorrect answer prediction.}
    \label{tab:ea}
    \begin{tabular}{lrr}
        \toprule
        \textbf{Main issues} & \textbf{CWQ} & \textbf{WebQSP} \\
        \midrule
        Insufficient AP & 20\% & 28\% \\
        EP construction errors & 6\% & 8\% \\
        EP ranking errors & 4\% & 20\% \\
        Non-entity evidence & 54\% & 30\% \\
        Answer reasoning errors & 8\% & 0\% \\
        Data quality & 8\% & 14\% \\
        \bottomrule
    \end{tabular}
\end{table}

Besides, our further analysis shows that unseen relation is a crucial issue for our implementation. About 2.8\% (98/3531) of CWQ test questions and 5\% (83/1639) of WebQSP test questions contain relations that never appear on the training questions. Our implementation experiences a sharp decline on these questions, as illustrated in table~\ref{tab:unseen}.

\begin{table}[htb]
    \centering
    \caption{The results (\%) on questions with and without unseen relations. All results are obtained with $K=100$.}
    \label{tab:unseen}
    \begin{tabular}{crrrrrr}
        \toprule
        \multirow{2}{*}{\textbf{Unseen Rel.}} & \multicolumn{3}{c}{\textbf{CWQ}} & \multicolumn{3}{c}{\textbf{WebQSP}} \\
        \cmidrule(lr){2-4} \cmidrule(lr){5-7}
        & C.R. & H@1 & F1 & C.R. & H@1 & F1 \\
        \midrule
        % 200 & Yes & 45.9 &	21.4 &	24.5 &	50.6 &	37.4 &	36.5  \\
        %  & No & 83.7 &	64.0 &	64.2 &	83.0 &	73.4 &	72.3  \\
        % \midrule[0.1pt]
        with & 36.7 &	17.4 &	19.7 &	49.4 &	36.1 &	35.4  \\
        w/o & 80.7 &	61.8 &	62.4 &	82.8 &	73.1 &	72.0  \\
        % \midrule[0.1pt]
        % 20 & Yes & 21.4 &	7.1 &	9.7 &	44.6 &	33.7 &	32.4  \\
        %  & No & 63.3 &	47.2 &	48.2 &	75.8 &	67.2 &	66.0  \\
        \bottomrule
    \end{tabular}
\end{table}

\section{Conclusion}
In this paper, we propose evidence pattern retrieval (EPR), which aims to improve the subgraph extraction of IR-KGQA methods by reducing noisy facts.
Our main contribution can be summarized as follows:
\begin{itemize}
    \item We propose the novel idea of evidence pattern, which refers to how necessary resources (entities and relations) are connected to support a knowledge graph node as an answer to a question. It enables the explicit modeling of structural dependencies during the subgraph extraction process of IR-KGQA.
    \item We propose an efficient implementation of EPR. It takes an evidence pattern as a combinations of the atomic adjacency patterns of resource pairs. We build a vector index for fast retrieval of the atomic patterns and propose an algorithm to construct evidence patterns. 
    \item We evaluate the EPR-based KGQA system with a rich experimental analysis. Our analysis demonstrates the importance of structural dependencies and EPR's ability to handle complex questions. 
\end{itemize}

Although EPR significantly enhances IR-KGQA methods in handling complex questions, there are still issues worth further exploration. In this paper, we implement atomic pattern retrieval using a BERT-based bi-encoder. Experimental results indicate that its performance is suboptimal in cases where the retrieval threshold is low (e.g., $\le$ 40) or where unseen relations are present in questions. It is necessary to explore solutions that perform better without significantly compromising retrieval efficiency. Our implementation on CWQ leads to a combinatorial explosion as the retrieval threshold increases, and it may require necessary optimizations for the brute-force enumeration of EP to improve efficiency. Exploring the combination of EPR and in-context learning to achieve state-of-the-art performance in a few-shot manner is worth investigating.
Besides, current IR-KGQA methods lack the ability to model numerical information. The possibility of modeling numerical features as pattern information to enhance the downstream answer reasoning methods' capability is also worth investigating.

\begin{acks}
This work is supported by the National Natural Science Foundation
of China (No. 62072224).
\end{acks}

\bibliographystyle{ACM-Reference-Format}
\bibliography{citation}


\begin{thebibliography}{38}

%%% the Web address.

\ifx \showCODEN    \undefined \def \showCODEN     #1{\unskip}     \fi
\ifx \showDOI      \undefined \def \showDOI       #1{#1}\fi
\ifx \showISBNx    \undefined \def \showISBNx     #1{\unskip}     \fi
\ifx \showISBNxiii \undefined \def \showISBNxiii  #1{\unskip}     \fi
\ifx \showISSN     \undefined \def \showISSN      #1{\unskip}     \fi
\ifx \showLCCN     \undefined \def \showLCCN      #1{\unskip}     \fi
\ifx \shownote     \undefined \def \shownote      #1{#1}          \fi
\ifx \showarticletitle \undefined \def \showarticletitle #1{#1}   \fi
\ifx \showURL      \undefined \def \showURL       {\relax}        \fi
% invisible to TeX
\providecommand\bibfield[2]{#2}
\providecommand\bibinfo[2]{#2}
\providecommand\natexlab[1]{#1}
\providecommand\showeprint[2][]{arXiv:#2}

\bibitem[Bao et~al\mbox{.}(2016)]%
        {bao-etal-2016-constraint}
\bibfield{author}{\bibinfo{person}{Junwei Bao}, \bibinfo{person}{Nan Duan}, \bibinfo{person}{Zhao Yan}, \bibinfo{person}{Ming Zhou}, {and} \bibinfo{person}{Tiejun Zhao}.} \bibinfo{year}{2016}\natexlab{}.
\newblock \showarticletitle{Constraint-Based Question Answering with Knowledge Graph}. In \bibinfo{booktitle}{\emph{Proceedings of {COLING} 2016, the 26th International Conference on Computational Linguistics: Technical Papers}}. \bibinfo{publisher}{The COLING 2016 Organizing Committee}, \bibinfo{address}{Osaka, Japan}, \bibinfo{pages}{2503--2514}.
\newblock
\urldef\tempurl%
\url{https://aclanthology.org/C16-1236}
\showURL{%
\tempurl}


\bibitem[Berant et~al\mbox{.}(2013)]%
        {webq}
\bibfield{author}{\bibinfo{person}{Jonathan Berant}, \bibinfo{person}{Andrew Chou}, \bibinfo{person}{Roy Frostig}, {and} \bibinfo{person}{Percy Liang}.} \bibinfo{year}{2013}\natexlab{}.
\newblock \showarticletitle{Semantic Parsing on {F}reebase from Question-Answer Pairs}. In \bibinfo{booktitle}{\emph{Proceedings of the 2013 Conference on Empirical Methods in Natural Language Processing}}. \bibinfo{publisher}{Association for Computational Linguistics}, \bibinfo{address}{Seattle, Washington, USA}, \bibinfo{pages}{1533--1544}.
\newblock
\urldef\tempurl%
\url{https://aclanthology.org/D13-1160}
\showURL{%
\tempurl}


\bibitem[Bollacker et~al\mbox{.}(2008)]%
        {freebase}
\bibfield{author}{\bibinfo{person}{Kurt Bollacker}, \bibinfo{person}{Colin Evans}, \bibinfo{person}{Praveen Paritosh}, \bibinfo{person}{Tim Sturge}, {and} \bibinfo{person}{Jamie Taylor}.} \bibinfo{year}{2008}\natexlab{}.
\newblock \showarticletitle{Freebase: A Collaboratively Created Graph Database for Structuring Human Knowledge}. In \bibinfo{booktitle}{\emph{Proceedings of the 2008 ACM SIGMOD International Conference on Management of Data}} (Vancouver, Canada) \emph{(\bibinfo{series}{SIGMOD '08})}. \bibinfo{publisher}{Association for Computing Machinery}, \bibinfo{address}{New York, NY, USA}, \bibinfo{pages}{1247–1250}.
\newblock
\showISBNx{9781605581026}
\urldef\tempurl%
\url{https://doi.org/10.1145/1376616.1376746}
\showDOI{\tempurl}


\bibitem[Bordes et~al\mbox{.}(2015)]%
        {simq}
\bibfield{author}{\bibinfo{person}{Antoine Bordes}, \bibinfo{person}{Nicolas Usunier}, \bibinfo{person}{Sumit Chopra}, {and} \bibinfo{person}{Jason Weston}.} \bibinfo{year}{2015}\natexlab{}.
\newblock \bibinfo{title}{Large-scale Simple Question Answering with Memory Networks}.
\newblock
\newblock
\showeprint[arxiv]{1506.02075}~[cs.LG]


\bibitem[Cai and Yates(2013)]%
        {free917}
\bibfield{author}{\bibinfo{person}{Qingqing Cai} {and} \bibinfo{person}{Alexander Yates}.} \bibinfo{year}{2013}\natexlab{}.
\newblock \showarticletitle{Large-scale Semantic Parsing via Schema Matching and Lexicon Extension}. In \bibinfo{booktitle}{\emph{Proceedings of the 51st Annual Meeting of the Association for Computational Linguistics (Volume 1: Long Papers)}}. \bibinfo{publisher}{Association for Computational Linguistics}, \bibinfo{address}{Sofia, Bulgaria}, \bibinfo{pages}{423--433}.
\newblock
\urldef\tempurl%
\url{https://aclanthology.org/P13-1042}
\showURL{%
\tempurl}


\bibitem[Cao et~al\mbox{.}(2022a)]%
        {kqapro}
\bibfield{author}{\bibinfo{person}{Shulin Cao}, \bibinfo{person}{Jiaxin Shi}, \bibinfo{person}{Liangming Pan}, \bibinfo{person}{Lunyiu Nie}, \bibinfo{person}{Yutong Xiang}, \bibinfo{person}{Lei Hou}, \bibinfo{person}{Juanzi Li}, \bibinfo{person}{Bin He}, {and} \bibinfo{person}{Hanwang Zhang}.} \bibinfo{year}{2022}\natexlab{a}.
\newblock \showarticletitle{{KQA} Pro: A Dataset with Explicit Compositional Programs for Complex Question Answering over Knowledge Base}. In \bibinfo{booktitle}{\emph{Proceedings of the 60th Annual Meeting of the Association for Computational Linguistics (Volume 1: Long Papers)}}. \bibinfo{publisher}{Association for Computational Linguistics}, \bibinfo{address}{Dublin, Ireland}, \bibinfo{pages}{6101--6119}.
\newblock
\urldef\tempurl%
\url{https://doi.org/10.18653/v1/2022.acl-long.422}
\showDOI{\tempurl}


\bibitem[Cao et~al\mbox{.}(2022b)]%
        {cao-etal-2022-program}
\bibfield{author}{\bibinfo{person}{Shulin Cao}, \bibinfo{person}{Jiaxin Shi}, \bibinfo{person}{Zijun Yao}, \bibinfo{person}{Xin Lv}, \bibinfo{person}{Jifan Yu}, \bibinfo{person}{Lei Hou}, \bibinfo{person}{Juanzi Li}, \bibinfo{person}{Zhiyuan Liu}, {and} \bibinfo{person}{Jinghui Xiao}.} \bibinfo{year}{2022}\natexlab{b}.
\newblock \showarticletitle{Program Transfer for Answering Complex Questions over Knowledge Bases}. In \bibinfo{booktitle}{\emph{Proceedings of the 60th Annual Meeting of the Association for Computational Linguistics (Volume 1: Long Papers)}}. \bibinfo{publisher}{Association for Computational Linguistics}, \bibinfo{address}{Dublin, Ireland}, \bibinfo{pages}{8128--8140}.
\newblock
\urldef\tempurl%
\url{https://doi.org/10.18653/v1/2022.acl-long.559}
\showDOI{\tempurl}


\bibitem[Chen et~al\mbox{.}(2018)]%
        {metaqa}
\bibfield{author}{\bibinfo{person}{Wenhu Chen}, \bibinfo{person}{Wenhan Xiong}, \bibinfo{person}{Xifeng Yan}, {and} \bibinfo{person}{William~Yang Wang}.} \bibinfo{year}{2018}\natexlab{}.
\newblock \showarticletitle{Variational Knowledge Graph Reasoning}. In \bibinfo{booktitle}{\emph{Proceedings of the 2018 Conference of the North {A}merican Chapter of the Association for Computational Linguistics: Human Language Technologies, Volume 1 (Long Papers)}}. \bibinfo{publisher}{Association for Computational Linguistics}, \bibinfo{address}{New Orleans, Louisiana}, \bibinfo{pages}{1823--1832}.
\newblock
\urldef\tempurl%
\url{https://doi.org/10.18653/v1/N18-1165}
\showDOI{\tempurl}


\bibitem[Das et~al\mbox{.}(2021)]%
        {cbr}
\bibfield{author}{\bibinfo{person}{Rajarshi Das}, \bibinfo{person}{Manzil Zaheer}, \bibinfo{person}{Dung Thai}, \bibinfo{person}{Ameya Godbole}, \bibinfo{person}{Ethan Perez}, \bibinfo{person}{Jay~Yoon Lee}, \bibinfo{person}{Lizhen Tan}, \bibinfo{person}{Lazaros Polymenakos}, {and} \bibinfo{person}{Andrew McCallum}.} \bibinfo{year}{2021}\natexlab{}.
\newblock \showarticletitle{Case-based Reasoning for Natural Language Queries over Knowledge Bases}. In \bibinfo{booktitle}{\emph{Proceedings of the 2021 Conference on Empirical Methods in Natural Language Processing}}. \bibinfo{publisher}{Association for Computational Linguistics}, \bibinfo{address}{Online and Punta Cana, Dominican Republic}, \bibinfo{pages}{9594--9611}.
\newblock
\urldef\tempurl%
\url{https://doi.org/10.18653/v1/2021.emnlp-main.755}
\showDOI{\tempurl}


\bibitem[Devlin et~al\mbox{.}(2019)]%
        {devlin-etal-2019-bert}
\bibfield{author}{\bibinfo{person}{Jacob Devlin}, \bibinfo{person}{Ming-Wei Chang}, \bibinfo{person}{Kenton Lee}, {and} \bibinfo{person}{Kristina Toutanova}.} \bibinfo{year}{2019}\natexlab{}.
\newblock \showarticletitle{{BERT}: Pre-training of Deep Bidirectional Transformers for Language Understanding}. In \bibinfo{booktitle}{\emph{Proceedings of the 2019 Conference of the North {A}merican Chapter of the Association for Computational Linguistics: Human Language Technologies, Volume 1 (Long and Short Papers)}}. \bibinfo{publisher}{Association for Computational Linguistics}, \bibinfo{address}{Minneapolis, Minnesota}, \bibinfo{pages}{4171--4186}.
\newblock
\urldef\tempurl%
\url{https://doi.org/10.18653/v1/N19-1423}
\showDOI{\tempurl}


\bibitem[Dong et~al\mbox{.}(2015)]%
        {dong-etal-2015-question}
\bibfield{author}{\bibinfo{person}{Li Dong}, \bibinfo{person}{Furu Wei}, \bibinfo{person}{Ming Zhou}, {and} \bibinfo{person}{Ke Xu}.} \bibinfo{year}{2015}\natexlab{}.
\newblock \showarticletitle{Question Answering over {F}reebase with Multi-Column Convolutional Neural Networks}. In \bibinfo{booktitle}{\emph{Proceedings of the 53rd Annual Meeting of the Association for Computational Linguistics and the 7th International Joint Conference on Natural Language Processing (Volume 1: Long Papers)}}. \bibinfo{publisher}{Association for Computational Linguistics}, \bibinfo{address}{Beijing, China}, \bibinfo{pages}{260--269}.
\newblock
\urldef\tempurl%
\url{https://doi.org/10.3115/v1/P15-1026}
\showDOI{\tempurl}


\bibitem[Dubey et~al\mbox{.}(2019)]%
        {lcquad2}
\bibfield{author}{\bibinfo{person}{Mohnish Dubey}, \bibinfo{person}{Debayan Banerjee}, \bibinfo{person}{Abdelrahman Abdelkawi}, {and} \bibinfo{person}{Jens Lehmann}.} \bibinfo{year}{2019}\natexlab{}.
\newblock \showarticletitle{LC-QuAD 2.0: A Large Dataset for Complex Question Answering over Wikidata and DBpedia}. In \bibinfo{booktitle}{\emph{The Semantic Web – ISWC 2019: 18th International Semantic Web Conference, Auckland, New Zealand, October 26–30, 2019, Proceedings, Part II}} (Auckland, New Zealand). \bibinfo{publisher}{Springer-Verlag}, \bibinfo{address}{Berlin, Heidelberg}, \bibinfo{pages}{69–78}.
\newblock
\showISBNx{978-3-030-30795-0}
\urldef\tempurl%
\url{https://doi.org/10.1007/978-3-030-30796-7_5}
\showDOI{\tempurl}


\bibitem[Gu et~al\mbox{.}(2023)]%
        {gu-etal-2023-dont}
\bibfield{author}{\bibinfo{person}{Yu Gu}, \bibinfo{person}{Xiang Deng}, {and} \bibinfo{person}{Yu Su}.} \bibinfo{year}{2023}\natexlab{}.
\newblock \showarticletitle{Don{'}t Generate, Discriminate: A Proposal for Grounding Language Models to Real-World Environments}. In \bibinfo{booktitle}{\emph{Proceedings of the 61st Annual Meeting of the Association for Computational Linguistics (Volume 1: Long Papers)}}. \bibinfo{publisher}{Association for Computational Linguistics}, \bibinfo{address}{Toronto, Canada}, \bibinfo{pages}{4928--4949}.
\newblock
\urldef\tempurl%
\url{https://doi.org/10.18653/v1/2023.acl-long.270}
\showDOI{\tempurl}


\bibitem[Gu et~al\mbox{.}(2021)]%
        {grailqa}
\bibfield{author}{\bibinfo{person}{Yu Gu}, \bibinfo{person}{Sue Kase}, \bibinfo{person}{Michelle Vanni}, \bibinfo{person}{Brian Sadler}, \bibinfo{person}{Percy Liang}, \bibinfo{person}{Xifeng Yan}, {and} \bibinfo{person}{Yu Su}.} \bibinfo{year}{2021}\natexlab{}.
\newblock \showarticletitle{Beyond I.I.D.: Three Levels of Generalization for Question Answering on Knowledge Bases}. In \bibinfo{booktitle}{\emph{Proceedings of the Web Conference 2021}} (Ljubljana, Slovenia) \emph{(\bibinfo{series}{WWW '21})}. \bibinfo{publisher}{Association for Computing Machinery}, \bibinfo{address}{New York, NY, USA}, \bibinfo{pages}{3477–3488}.
\newblock
\showISBNx{9781450383127}
\urldef\tempurl%
\url{https://doi.org/10.1145/3442381.3449992}
\showDOI{\tempurl}


\bibitem[He et~al\mbox{.}(2021)]%
        {nsm-h}
\bibfield{author}{\bibinfo{person}{Gaole He}, \bibinfo{person}{Yunshi Lan}, \bibinfo{person}{Jing Jiang}, \bibinfo{person}{Wayne~Xin Zhao}, {and} \bibinfo{person}{Ji-Rong Wen}.} \bibinfo{year}{2021}\natexlab{}.
\newblock \showarticletitle{Improving Multi-Hop Knowledge Base Question Answering by Learning Intermediate Supervision Signals}. In \bibinfo{booktitle}{\emph{Proceedings of the 14th ACM International Conference on Web Search and Data Mining}} (Virtual Event, Israel) \emph{(\bibinfo{series}{WSDM '21})}. \bibinfo{publisher}{Association for Computing Machinery}, \bibinfo{address}{New York, NY, USA}, \bibinfo{pages}{553–561}.
\newblock
\showISBNx{9781450382977}
\urldef\tempurl%
\url{https://doi.org/10.1145/3437963.3441753}
\showDOI{\tempurl}


\bibitem[Hu et~al\mbox{.}(2018)]%
        {hu-etal-2018-state}
\bibfield{author}{\bibinfo{person}{Sen Hu}, \bibinfo{person}{Lei Zou}, {and} \bibinfo{person}{Xinbo Zhang}.} \bibinfo{year}{2018}\natexlab{}.
\newblock \showarticletitle{A State-transition Framework to Answer Complex Questions over Knowledge Base}. In \bibinfo{booktitle}{\emph{Proceedings of the 2018 Conference on Empirical Methods in Natural Language Processing}}. \bibinfo{publisher}{Association for Computational Linguistics}, \bibinfo{address}{Brussels, Belgium}, \bibinfo{pages}{2098--2108}.
\newblock
\urldef\tempurl%
\url{https://doi.org/10.18653/v1/D18-1234}
\showDOI{\tempurl}


\bibitem[Hu et~al\mbox{.}(2022)]%
        {hu-etal-2022-logical}
\bibfield{author}{\bibinfo{person}{Xixin Hu}, \bibinfo{person}{Xuan Wu}, \bibinfo{person}{Yiheng Shu}, {and} \bibinfo{person}{Yuzhong Qu}.} \bibinfo{year}{2022}\natexlab{}.
\newblock \showarticletitle{Logical Form Generation via Multi-task Learning for Complex Question Answering over Knowledge Bases}. In \bibinfo{booktitle}{\emph{Proceedings of the 29th International Conference on Computational Linguistics}}. \bibinfo{publisher}{International Committee on Computational Linguistics}, \bibinfo{address}{Gyeongju, Republic of Korea}, \bibinfo{pages}{1687--1696}.
\newblock
\urldef\tempurl%
\url{https://aclanthology.org/2022.coling-1.145}
\showURL{%
\tempurl}


\bibitem[Jiang et~al\mbox{.}(2023)]%
        {unikgqa}
\bibfield{author}{\bibinfo{person}{Jinhao Jiang}, \bibinfo{person}{Kun Zhou}, \bibinfo{person}{Xin Zhao}, {and} \bibinfo{person}{Ji-Rong Wen}.} \bibinfo{year}{2023}\natexlab{}.
\newblock \showarticletitle{Uni{KGQA}: Unified Retrieval and Reasoning for Solving Multi-hop Question Answering Over Knowledge Graph}. In \bibinfo{booktitle}{\emph{The Eleventh International Conference on Learning Representations}}.
\newblock
\urldef\tempurl%
\url{https://openreview.net/forum?id=Z63RvyAZ2Vh}
\showURL{%
\tempurl}


\bibitem[Johnson et~al\mbox{.}(2021)]%
        {faiss}
\bibfield{author}{\bibinfo{person}{Jeff Johnson}, \bibinfo{person}{Matthijs Douze}, {and} \bibinfo{person}{Hervé Jégou}.} \bibinfo{year}{2021}\natexlab{}.
\newblock \showarticletitle{Billion-Scale Similarity Search with GPUs}.
\newblock \bibinfo{journal}{\emph{IEEE Transactions on Big Data}} \bibinfo{volume}{7}, \bibinfo{number}{3} (\bibinfo{year}{2021}), \bibinfo{pages}{535--547}.
\newblock
\urldef\tempurl%
\url{https://doi.org/10.1109/TBDATA.2019.2921572}
\showDOI{\tempurl}


\bibitem[Karpukhin et~al\mbox{.}(2020)]%
        {dpr}
\bibfield{author}{\bibinfo{person}{Vladimir Karpukhin}, \bibinfo{person}{Barlas Oguz}, \bibinfo{person}{Sewon Min}, \bibinfo{person}{Patrick Lewis}, \bibinfo{person}{Ledell Wu}, \bibinfo{person}{Sergey Edunov}, \bibinfo{person}{Danqi Chen}, {and} \bibinfo{person}{Wen-tau Yih}.} \bibinfo{year}{2020}\natexlab{}.
\newblock \showarticletitle{Dense Passage Retrieval for Open-Domain Question Answering}. In \bibinfo{booktitle}{\emph{Proceedings of the 2020 Conference on Empirical Methods in Natural Language Processing (EMNLP)}}. \bibinfo{publisher}{Association for Computational Linguistics}, \bibinfo{address}{Online}, \bibinfo{pages}{6769--6781}.
\newblock
\urldef\tempurl%
\url{https://doi.org/10.18653/v1/2020.emnlp-main.550}
\showDOI{\tempurl}


\bibitem[Lan et~al\mbox{.}(2023)]%
        {survey}
\bibfield{author}{\bibinfo{person}{Yunshi Lan}, \bibinfo{person}{Gaole He}, \bibinfo{person}{Jinhao Jiang}, \bibinfo{person}{Jing Jiang}, \bibinfo{person}{Wayne~Xin Zhao}, {and} \bibinfo{person}{Ji-Rong Wen}.} \bibinfo{year}{2023}\natexlab{}.
\newblock \showarticletitle{Complex Knowledge Base Question Answering: A Survey}.
\newblock \bibinfo{journal}{\emph{IEEE Transactions on Knowledge and Data Engineering}} \bibinfo{volume}{35}, \bibinfo{number}{11} (\bibinfo{year}{2023}), \bibinfo{pages}{11196--11215}.
\newblock
\urldef\tempurl%
\url{https://doi.org/10.1109/TKDE.2022.3223858}
\showDOI{\tempurl}


\bibitem[Li et~al\mbox{.}(2023b)]%
        {li-etal-2023-shot}
\bibfield{author}{\bibinfo{person}{Tianle Li}, \bibinfo{person}{Xueguang Ma}, \bibinfo{person}{Alex Zhuang}, \bibinfo{person}{Yu Gu}, \bibinfo{person}{Yu Su}, {and} \bibinfo{person}{Wenhu Chen}.} \bibinfo{year}{2023}\natexlab{b}.
\newblock \showarticletitle{Few-shot In-context Learning on Knowledge Base Question Answering}. In \bibinfo{booktitle}{\emph{Proceedings of the 61st Annual Meeting of the Association for Computational Linguistics (Volume 1: Long Papers)}}. \bibinfo{publisher}{Association for Computational Linguistics}, \bibinfo{address}{Toronto, Canada}, \bibinfo{pages}{6966--6980}.
\newblock
\urldef\tempurl%
\url{https://doi.org/10.18653/v1/2023.acl-long.385}
\showDOI{\tempurl}


\bibitem[Li et~al\mbox{.}(2023a)]%
        {li2023flexkbqa}
\bibfield{author}{\bibinfo{person}{Zhenyu Li}, \bibinfo{person}{Sunqi Fan}, \bibinfo{person}{Yu Gu}, \bibinfo{person}{Xiuxing Li}, \bibinfo{person}{Zhichao Duan}, \bibinfo{person}{Bowen Dong}, \bibinfo{person}{Ning Liu}, {and} \bibinfo{person}{Jianyong Wang}.} \bibinfo{year}{2023}\natexlab{a}.
\newblock \bibinfo{title}{FlexKBQA: A Flexible LLM-Powered Framework for Few-Shot Knowledge Base Question Answering}.
\newblock
\newblock
\showeprint[arxiv]{2308.12060}~[cs.CL]


\bibitem[Miller et~al\mbox{.}(2016)]%
        {kvmem}
\bibfield{author}{\bibinfo{person}{Alexander Miller}, \bibinfo{person}{Adam Fisch}, \bibinfo{person}{Jesse Dodge}, \bibinfo{person}{Amir-Hossein Karimi}, \bibinfo{person}{Antoine Bordes}, {and} \bibinfo{person}{Jason Weston}.} \bibinfo{year}{2016}\natexlab{}.
\newblock \showarticletitle{Key-Value Memory Networks for Directly Reading Documents}. In \bibinfo{booktitle}{\emph{Proceedings of the 2016 Conference on Empirical Methods in Natural Language Processing}}. \bibinfo{publisher}{Association for Computational Linguistics}, \bibinfo{address}{Austin, Texas}, \bibinfo{pages}{1400--1409}.
\newblock


\bibitem[Nie et~al\mbox{.}(2023)]%
        {nie2023codestyle}
\bibfield{author}{\bibinfo{person}{Zhijie Nie}, \bibinfo{person}{Richong Zhang}, \bibinfo{person}{Zhongyuan Wang}, {and} \bibinfo{person}{Xudong Liu}.} \bibinfo{year}{2023}\natexlab{}.
\newblock \bibinfo{title}{Code-Style In-Context Learning for Knowledge-Based Question Answering}.
\newblock
\newblock
\showeprint[arxiv]{2309.04695}~[cs.CL]


\bibitem[Saxena et~al\mbox{.}(2020)]%
        {embedded_kgqa}
\bibfield{author}{\bibinfo{person}{Apoorv Saxena}, \bibinfo{person}{Aditay Tripathi}, {and} \bibinfo{person}{Partha Talukdar}.} \bibinfo{year}{2020}\natexlab{}.
\newblock \showarticletitle{Improving Multi-hop Question Answering over Knowledge Graphs using Knowledge Base Embeddings}. In \bibinfo{booktitle}{\emph{Proceedings of the 58th Annual Meeting of the Association for Computational Linguistics}}. \bibinfo{publisher}{Association for Computational Linguistics}, \bibinfo{address}{Online}, \bibinfo{pages}{4498--4507}.
\newblock


\bibitem[Shu et~al\mbox{.}(2022)]%
        {tiara}
\bibfield{author}{\bibinfo{person}{Yiheng Shu}, \bibinfo{person}{Zhiwei Yu}, \bibinfo{person}{Yuhan Li}, \bibinfo{person}{B{\"o}rje Karlsson}, \bibinfo{person}{Tingting Ma}, \bibinfo{person}{Yuzhong Qu}, {and} \bibinfo{person}{Chin-Yew Lin}.} \bibinfo{year}{2022}\natexlab{}.
\newblock \showarticletitle{{TIARA}: Multi-grained Retrieval for Robust Question Answering over Large Knowledge Base}. In \bibinfo{booktitle}{\emph{Proceedings of the 2022 Conference on Empirical Methods in Natural Language Processing}}. \bibinfo{publisher}{Association for Computational Linguistics}, \bibinfo{address}{Abu Dhabi, United Arab Emirates}, \bibinfo{pages}{8108--8121}.
\newblock
\urldef\tempurl%
\url{https://aclanthology.org/2022.emnlp-main.555}
\showURL{%
\tempurl}


\bibitem[Sun et~al\mbox{.}(2019)]%
        {pullnet}
\bibfield{author}{\bibinfo{person}{Haitian Sun}, \bibinfo{person}{Tania Bedrax-Weiss}, {and} \bibinfo{person}{William Cohen}.} \bibinfo{year}{2019}\natexlab{}.
\newblock \showarticletitle{{P}ull{N}et: Open Domain Question Answering with Iterative Retrieval on Knowledge Bases and Text}. In \bibinfo{booktitle}{\emph{Proceedings of the 2019 Conference on Empirical Methods in Natural Language Processing and the 9th International Joint Conference on Natural Language Processing (EMNLP-IJCNLP)}}. \bibinfo{publisher}{Association for Computational Linguistics}, \bibinfo{address}{Hong Kong, China}, \bibinfo{pages}{2380--2390}.
\newblock


\bibitem[Sun et~al\mbox{.}(2018)]%
        {graftnet}
\bibfield{author}{\bibinfo{person}{Haitian Sun}, \bibinfo{person}{Bhuwan Dhingra}, \bibinfo{person}{Manzil Zaheer}, \bibinfo{person}{Kathryn Mazaitis}, \bibinfo{person}{Ruslan Salakhutdinov}, {and} \bibinfo{person}{William Cohen}.} \bibinfo{year}{2018}\natexlab{}.
\newblock \showarticletitle{Open Domain Question Answering Using Early Fusion of Knowledge Bases and Text}. In \bibinfo{booktitle}{\emph{Proceedings of the 2018 Conference on Empirical Methods in Natural Language Processing}}. \bibinfo{publisher}{Association for Computational Linguistics}, \bibinfo{address}{Brussels, Belgium}, \bibinfo{pages}{4231--4242}.
\newblock


\bibitem[Talmor and Berant(2018)]%
        {cwq}
\bibfield{author}{\bibinfo{person}{Alon Talmor} {and} \bibinfo{person}{Jonathan Berant}.} \bibinfo{year}{2018}\natexlab{}.
\newblock \showarticletitle{The Web as a Knowledge-Base for Answering Complex Questions}. In \bibinfo{booktitle}{\emph{Proceedings of the 2018 Conference of the North {A}merican Chapter of the Association for Computational Linguistics: Human Language Technologies, Volume 1 (Long Papers)}}. \bibinfo{publisher}{Association for Computational Linguistics}, \bibinfo{address}{New Orleans, Louisiana}, \bibinfo{pages}{641--651}.
\newblock


\bibitem[Trivedi et~al\mbox{.}(2017)]%
        {lcquad}
\bibfield{author}{\bibinfo{person}{Priyansh Trivedi}, \bibinfo{person}{Gaurav Maheshwari}, \bibinfo{person}{Mohnish Dubey}, {and} \bibinfo{person}{Jens Lehmann}.} \bibinfo{year}{2017}\natexlab{}.
\newblock \showarticletitle{LC-QuAD: A Corpus for Complex Question Answering over Knowledge Graphs}. In \bibinfo{booktitle}{\emph{The Semantic Web -- ISWC 2017}}, \bibfield{editor}{\bibinfo{person}{Claudia d'Amato}, \bibinfo{person}{Miriam Fernandez}, \bibinfo{person}{Valentina Tamma}, \bibinfo{person}{Freddy Lecue}, \bibinfo{person}{Philippe Cudr{\'e}-Mauroux}, \bibinfo{person}{Juan Sequeda}, \bibinfo{person}{Christoph Lange}, {and} \bibinfo{person}{Jeff Heflin}} (Eds.). \bibinfo{publisher}{Springer International Publishing}, \bibinfo{address}{Cham}, \bibinfo{pages}{210--218}.
\newblock
\showISBNx{978-3-319-68204-4}


\bibitem[Unger et~al\mbox{.}(2014)]%
        {qald}
\bibfield{author}{\bibinfo{person}{Christina Unger}, \bibinfo{person}{Andr{\'e} Freitas}, {and} \bibinfo{person}{Philipp Cimiano}.} \bibinfo{year}{2014}\natexlab{}.
\newblock \bibinfo{booktitle}{\emph{An Introduction to Question Answering over Linked Data}}.
\newblock \bibinfo{publisher}{Springer International Publishing}, \bibinfo{address}{Cham}, \bibinfo{pages}{100--140}.
\newblock
\showISBNx{978-3-319-10587-1}
\urldef\tempurl%
\url{https://doi.org/10.1007/978-3-319-10587-1_2}
\showDOI{\tempurl}


\bibitem[Wang et~al\mbox{.}(2023)]%
        {wang2023knowledgedriven}
\bibfield{author}{\bibinfo{person}{Keheng Wang}, \bibinfo{person}{Feiyu Duan}, \bibinfo{person}{Sirui Wang}, \bibinfo{person}{Peiguang Li}, \bibinfo{person}{Yunsen Xian}, \bibinfo{person}{Chuantao Yin}, \bibinfo{person}{Wenge Rong}, {and} \bibinfo{person}{Zhang Xiong}.} \bibinfo{year}{2023}\natexlab{}.
\newblock \bibinfo{title}{Knowledge-Driven CoT: Exploring Faithful Reasoning in LLMs for Knowledge-intensive Question Answering}.
\newblock
\newblock
\showeprint[arxiv]{2308.13259}~[cs.CL]


\bibitem[Xu et~al\mbox{.}(2016)]%
        {xu-etal-2016-question}
\bibfield{author}{\bibinfo{person}{Kun Xu}, \bibinfo{person}{Siva Reddy}, \bibinfo{person}{Yansong Feng}, \bibinfo{person}{Songfang Huang}, {and} \bibinfo{person}{Dongyan Zhao}.} \bibinfo{year}{2016}\natexlab{}.
\newblock \showarticletitle{Question Answering on {F}reebase via Relation Extraction and Textual Evidence}. In \bibinfo{booktitle}{\emph{Proceedings of the 54th Annual Meeting of the Association for Computational Linguistics (Volume 1: Long Papers)}}. \bibinfo{publisher}{Association for Computational Linguistics}, \bibinfo{address}{Berlin, Germany}, \bibinfo{pages}{2326--2336}.
\newblock
\urldef\tempurl%
\url{https://doi.org/10.18653/v1/P16-1220}
\showDOI{\tempurl}


\bibitem[Ye et~al\mbox{.}(2022)]%
        {rngkbqa}
\bibfield{author}{\bibinfo{person}{Xi Ye}, \bibinfo{person}{Semih Yavuz}, \bibinfo{person}{Kazuma Hashimoto}, \bibinfo{person}{Yingbo Zhou}, {and} \bibinfo{person}{Caiming Xiong}.} \bibinfo{year}{2022}\natexlab{}.
\newblock \showarticletitle{{RNG}-{KBQA}: Generation Augmented Iterative Ranking for Knowledge Base Question Answering}. In \bibinfo{booktitle}{\emph{Proceedings of the 60th Annual Meeting of the Association for Computational Linguistics (Volume 1: Long Papers)}}. \bibinfo{publisher}{Association for Computational Linguistics}, \bibinfo{address}{Dublin, Ireland}, \bibinfo{pages}{6032--6043}.
\newblock
\urldef\tempurl%
\url{https://doi.org/10.18653/v1/2022.acl-long.417}
\showDOI{\tempurl}


\bibitem[Yih et~al\mbox{.}(2016)]%
        {webqsp}
\bibfield{author}{\bibinfo{person}{Wen-tau Yih}, \bibinfo{person}{Matthew Richardson}, \bibinfo{person}{Chris Meek}, \bibinfo{person}{Ming-Wei Chang}, {and} \bibinfo{person}{Jina Suh}.} \bibinfo{year}{2016}\natexlab{}.
\newblock \showarticletitle{The Value of Semantic Parse Labeling for Knowledge Base Question Answering}. In \bibinfo{booktitle}{\emph{Proceedings of the 54th Annual Meeting of the Association for Computational Linguistics (Volume 2: Short Papers)}}. \bibinfo{publisher}{Association for Computational Linguistics}, \bibinfo{address}{Berlin, Germany}, \bibinfo{pages}{201--206}.
\newblock
\urldef\tempurl%
\url{https://doi.org/10.18653/v1/P16-2033}
\showDOI{\tempurl}


\bibitem[Yu et~al\mbox{.}(2023)]%
        {yu2023decaf}
\bibfield{author}{\bibinfo{person}{Donghan Yu}, \bibinfo{person}{Sheng Zhang}, \bibinfo{person}{Patrick Ng}, \bibinfo{person}{Henghui Zhu}, \bibinfo{person}{Alexander~Hanbo Li}, \bibinfo{person}{Jun Wang}, \bibinfo{person}{Yiqun Hu}, \bibinfo{person}{William~Yang Wang}, \bibinfo{person}{Zhiguo Wang}, {and} \bibinfo{person}{Bing Xiang}.} \bibinfo{year}{2023}\natexlab{}.
\newblock \showarticletitle{Dec{AF}: Joint Decoding of Answers and Logical Forms for Question Answering over Knowledge Bases}. In \bibinfo{booktitle}{\emph{The Eleventh International Conference on Learning Representations}}.
\newblock
\urldef\tempurl%
\url{https://openreview.net/forum?id=XHc5zRPxqV9}
\showURL{%
\tempurl}


\bibitem[Zhang et~al\mbox{.}(2022)]%
        {sr}
\bibfield{author}{\bibinfo{person}{Jing Zhang}, \bibinfo{person}{Xiaokang Zhang}, \bibinfo{person}{Jifan Yu}, \bibinfo{person}{Jian Tang}, \bibinfo{person}{Jie Tang}, \bibinfo{person}{Cuiping Li}, {and} \bibinfo{person}{Hong Chen}.} \bibinfo{year}{2022}\natexlab{}.
\newblock \showarticletitle{Subgraph Retrieval Enhanced Model for Multi-hop Knowledge Base Question Answering}. In \bibinfo{booktitle}{\emph{Proceedings of the 60th Annual Meeting of the Association for Computational Linguistics (Volume 1: Long Papers)}}. \bibinfo{publisher}{Association for Computational Linguistics}, \bibinfo{address}{Dublin, Ireland}, \bibinfo{pages}{5773--5784}.
\newblock
\urldef\tempurl%
\url{https://doi.org/10.18653/v1/2022.acl-long.396}
\showDOI{\tempurl}


\end{thebibliography}

% \appendix

\end{document}